\documentclass[a4paper]{article}

\usepackage{INTERSPEECH2019}
\usepackage{url}
\usepackage{multirow}

\title{Disfluencies and Human Speech Transcription Errors}
\name{Vicky Zayats$^1$, Trang Tran$^1$, Richard Wright$^2$, Courtney Mansfield$^2$, Mari Ostendorf$^1$, }
\address{
  $^1$Electrical \& Computer Engineering Department, University of Washington\\
  $^2$Linguistics Department, University of Washington}
\email{\{vzayats,ttmt001,rawright,coman8,ostendor\}@uw.edu}

\begin{document}

\maketitle
\begin{abstract}
This paper explores contexts associated with errors in transcription of spontaneous speech, shedding light on human perception of disfluencies and other conversational speech phenomena. A new version of the Switchboard corpus is provided with disfluency annotations for careful speech transcripts, together with results showing the impact of transcription errors on evaluation of automatic disfluency detection.
\end{abstract}

\section{Introduction}
Human errors in transcribing spontaneous speech provide insights into perception of speech and how humans process spoken language. In this study, we investigate mistranscriptions of spontaneous speech, particularly the co-occurrence of errors (or misperceptions) with disfluencies and coarse-grained word classes.
The motivation is twofold. First, the analysis of errors can shed light on the function(s) of disfluencies and other spontaneous speech phenomena in conversation. Second, given a better understanding of where errors occur improves our ability to interpret results from assessing automatic algorithms against human transcriptions and associated findings. 

Much of the work on misperception has been on read speech in a laboratory environment, where lexical items and grammatical structure can be controlled for. 
The two transcriptions of Switchboard provide an opportunity to investigate the properties of spontaneous speech associated with perception errors.
Our work on spontaneous speech provides new insights by averaging statistics over many instances of different word categories. This allows us to look at phenomena such as disfluencies, where we find higher rates of misperception (assuming mistranscription indicates misperception). 

The study is based on the Switchboard Corpus of conversational telephone speech \cite{godfrey1992switchboard,Godfrey1993}, which is a large collection of conversational telephone speech that has been annotated for a number of different types of linguistic structure. Importantly for this study, there are two sets of human transcriptions. The second represents a careful correction of the first, but much of the linguistic annotation is based on an earlier release.

We pose two hypotheses. First, we expect that words which carry more information are less likely to be misperceived, since they tend to be more clearly articulated.  Second, we anticipate that disfluent regions will be disproportionately associated with misperceptions, particularly repetition disfluencies. 
In order to carry out the second analysis, it is necessary to revise the existing disfluency annotation for the more careful transcriptions. We describe an automatic procedure that gives high quality results and make the resulting annotations publicly available.

The paper makes three main contributions. First, we present distributional analyses of the location of transcription errors that confirm our hypotheses, but also point to informal words used in conversation as a high error category.
Second, to support the analysis, a new version 
of disfluency annotated Switchboard is released. Finally, experiments on disfluency detection show that transcription errors impact performance estimates mainly for repair disfluencies.

\section{Human Speech Transcription}

With the fast improvements in ASR system performance, there have been multiple studies that evaluate human transcription accuracy on a variety of datasets. One of the early studies on human transcription \cite{lippmann1997speech} reports error rates across multiple corpora with different level of difficulty and vocabulary sizes ranging from 10 to 65k words. The human transcription rates in this study varies from 0.1\% (on transcribing 10 digits) to 7.4\% (keyword spotting) depending on the task and corpora. The study also estimates the error rate for the Switchboard corpus to be 4\%, although the reference attributed this number to ``personal communication." Later, independent studies \cite{xiong2016achieving,xiong2017toward,saon2017english} re-evaluated this number using professional transcribers, reporting error rates of 5.1-5.9\% and 6.8-11.3\% on subsets of the Switchboard (21k words) and CallHome (22k words) corpora, respectively, which are part of NIST 2000 CTS evaluation set. 
The differences in error rate correspond to how careful the transcribers are, with a quality checking pass improving the average error rate by 5-20\% \cite{saon2017english}. 
Another study by LDC on quality of the transcription \cite{glenn2010transcription} reveals huge differences in the transcription error rate between very careful (4.1-4.5\%) and quick transcription (9.6\%) evaluated on the RT-03 evaluation set, which contains subsets of Fisher and Switchboard datasets (76k words in total). The authors report that with very careful transcriptions 95\% of annotation discrepancies between multiple transcribers are ``judgment calls'' due to contractions, rapid or difficult speech, or disfluencies. The authors also notice that with the quick transcriptions, the regions of disfluency are by far the most prevalent contributors to transcriber disagreement across the different languages used in the study.
Transcription errors in genres other than conversational telephone speech (broadcast news, broadcast conversation, interviews, and meetings) are approximated at rates 1.3-6.3\% in English, 6.1-9.5\% in Chinese and 3.1-8.3\% in Arabic.

When trying to analyze and compare human and machine errors \cite{saon2017english,stolcke2017comparing} using NIST 2000 CTS dataset, both papers report function and backchannel words being dominant word categories labeled as errors, though due to a limited data size the statistics on frequent word associated with errors can be unreliable.
For example, the most common insertion (by human transcribers) is token "i" with only 10 occurrences  \cite{stolcke2002srilm}. 
In comparison, our work explores transcription errors using a large scale dataset (1.3M words), which allows identification of more reliable patterns and more fine-grained analysis of error contexts.

\section{Data}


Switchboard I -- Release 2 \cite{godfrey1992switchboard,Godfrey1993} is a collection of about 2,400 telephone conversations between strangers, of which 1126 conversations were hand-annotated with disfluencies as part of the Penn Treebank Release 3 dataset \cite{treebank3}. Because human transcribers are imperfect, the original transcripts contained errors, some of which were corrected in the Treebank release. Mississippi State University researchers ran a clean-up project which hand-corrected conversations and produced more accurate word alignments, indicating the type of errors (missed, extra, or substituted) \cite{Deshmukh1998}, 
which we will refer to as MsState transcriptions. While they added words associated with disfluencies, they did not re-annotate disfluency (or parse) structure. 

The MsState transcription guidelines were designed for higher consistency, so some of the transcription ``errors'' reflect a difference in transcription guidelines. In particular, the MsState transcription guidelines differ from the original guidelines in asking the transcriber to more faithfully represent the spoken version, e.g. by allowing more variants of words (e.g. ``naw'' for ``no,'' ``gonna'' for ``going to,'' and ``um-hum'' as well as ``uh-huh''), encouraging (rather than discouraging) use of contractions, having explicit conventions for pronunciation variants and mispronunciations, and in the form for transcribing word fragments (``w[ent]-'' vs. ``w-'').  In addition, the conventions for handling word fragments differs in terms of conventions for using ``I-'' in a repetition, and in asking the transcriber to include the fragment even if they do not know what the speaker intended, which leads to a higher rate of word fragments in the MsState transcripts. These convention differences can inflate the substitution error rate, and some are therefore ignored in our analyses.

The MsState project provided word-level alignments of the revised transcription to the Treebank transcripts, where each difference is labeled as an insertion, deletion or substitution.
In discussing the transcription differences, we refer to words marked as deleted in Treebank compared to the MsState transcript as ``missed,'' and those that are inserted in the Treebank transcriptions are called ``hallucinated.'' 

The original work documenting the segmentation and transcription correction effort \cite{Deshmukh1998} states that the human transcription word error rate is reduced from approximately 10\% to 2\%.
Our analysis on the disfluency-annotated subset shows a word difference rate of 5\%, 
using the standard error rate calculation (insertions + deletions + substitutions/total number of words in the MsState transcript), with 2.4\% associated with substitutions. 
(If contractions are split, as in many language processing studies, the word difference rate is 5.2\%, with 2.6\% substitutions.)
The smaller error rate may be due in part to the fact that we did not count differences due to transcription conventions as errors, e.g. (``i-'' vs ``I'', ``uh-hum'' vs ``uh-huh'') and differences associated with the CONT transcription, used e.g.\ for acronyms. It is also consistent with other studies describe in the previous section. In addition, if the 10\% error rate is based on the original release of the Switchboard transcripts, then there would also be a difference related to the fact that there were some corrections in the Treebank release. 

This error rate includes many word fragments: the MsState transcripts contain roughly 2.5 
times more word fragments than the Treebank transcripts. Ignoring the single-phone word fragments, which represent roughly 75\% 
of the fragments added in the retranscription, the error rate is 4.7\% (5.0\% with split contractions). As expected, word fragments are more often missed than inserted: 12.4\% vs.\ 7.2\%, respectively (11.8\% vs.\ 7.0\% for split contractions). 


\subsection{Automatic Mapping of Disfluency Annotations}

A goal of this research was to align the MsState speech transcripts (for which there are more careful transcripts and good time alignments) with disfluencies that had been hand-annotated on an earlier (less faithful) version of the transcripts. 
In order to transfer the disfluency annotations to the MsState transcripts, we used a multi-step process that leverages our previous work on transcript differences \cite{Zayats2015} and avoids a costly hand-annotation process. 
Each word in the original Treebank annotation was associated with a disfluency label based on a begin-inside-outside (BIO) tagging scheme that accounted for both reparandum and correction spans following \cite{Zayats2016}.
Using the MsState alignments, we inserted, deleted and substituted words in the disfluency transcripts. For each inserted or substituted words and the window of $\pm 2$ neighbors, two
additional (temporary) labels were used: 'D' for representing any state that correspond to being part of disfluency (either reparandum or repair), and 'A' which would allow any state. In addition, we used the 'D' state for words surrounding deletions that were originally annotated as disfluencies, assigned non-disfluency state 'O' for words surrounding insertions that were originally annotated as non-disfluencies and 'A' for all the rest. Then, we ran automatic disfluency detection with integer linear programming constraints \cite{Georgila2009,Zayats2016} for assigning BIO labels to the words associated with 'A' and 'D' labels. This approach allowed us to identify and add missed disfluencies and remove hallucinated disfluencies. We refer to the resulting annotations as ``silver'' annotations, since they are strongly constrained by hand annotations. While some errors are introduced in this process, most transcription errors are short and isolated, so the constraints of labels on neighboring words are reasonably strong.
As discussed in the next section, analysis of a subset of the test set shows that the mapped labels are quite good. We also show that the automatically corrected data used in training a disfluency detection model (vs. the Treebank annotations) leads to better performance.



\begin{table}
  \centering
  \label{tab:annotation_examples}
  \begin{tabular}{|l|l|} \hline
    Transcript & Annotation \\ \hline
    Treebank & and also the [ whole + whole ] thing \\ \hline
    MsState & and also the [ whole \{DEL the \} + whole ] thing \\ \hline
    Mapped & and also [ the whole +  the whole ] thing \\ \hline
  \end{tabular}
\vskip 0.1in
  \caption{Example of transferring disfluency annotation from Treebank to MsState transcripts.\label{tab:disfl_map}}
    
\end{table}


\subsection{Quality of Automatic Mappings}
\label{sec:quality}


To assess the quality of the automatically mapped data and the improvements in the mapping associated with the new disfluency model, we initially selected 100 test 
sentences\footnote{For simplicity, we use the term ``sentence" rather than sentence-like unit or slash unit (SU), which are sometimes used to describe these segmentations since they do not always have the complete grammatical structure of written sentences. The term ``slash unit'' comes from the use of ``./'' and ``-/'' to mark the boundaries of complete and interrupted units. } 
to hand correct for disfluencies. In annotating these, we observed that there were sentence segmentation alignment errors in some of the cases for which missed words occurred at sentence boundaries. We therefore selected additional sentences to hand annotate, separately computing statistics for those that included missed words at boundaries and those that did not.

Roughly 15\% of the 100 test sentences appeared to have missed words associated with a segmentation boundary. 
Because the available alignments for between the MsState and Treebank transcripts did not preserve sentence boundaries, a mechanism is needed to align deleted (missed) words to sentences. Initially, we arbitrarily assigned these to the sentence following the boundary. Examining 63 sentence pairs for possible boundary alignment problems, we found that 27\% involved assignment errors. 
Of the 17 sentences that had errors, two simple reassignment rules related to unintelligible regions and backchannels addressed 13 cases.
Using these rules, the full data set was reprocessed, and the problems seemed to be minimal in the subsequent hand annotation effort.


After automatic refinement of sentence segmentation, we annotated additional sentences, resulting in 453 sentences in total.
For this set, we compared the mapped silver annotations to the gold annotations. The reparandum labels are very good with
F-score of 90.1 (90.1 precision, 90.1 recall). The silver interruption points have F1 90.6 (89.4 precision, 91.8 recall). It is difficult to use standard disfluency detection scoring to characterize the quality of the original Treebank transcriptions because of the word sequence differences. 
However, we can easily assess the detected interruption points, and we find that the original Treebank annotations have much lower quality, with F1 79.7 (precision 88.9, recall 72.2). 
Most of the difference is in recall, consistent with the hypothesis that many transcription errors are in disfluent (reparandum) regions.


\section{Analysis of Transcription Errors}

In the analyses below, we distinguish between two types of misperceptions: i)~a word that appears in the careful MsState transcript but does not appear in the Treebank transcript, referred to as a `miss,' and ii)~a word that appears in the Treebank transcript but not in the MsState transcript, referred to as a `hallucination.' For differences in the two transcripts that correspond to a substitution error, the word that is in the Treebank transcript is counted as a hallucination, and the word that is in the MsState transcript is counted as a miss.

\subsection{Word Category}
We first looked at misperceptions depending on word category, classifying words as lexical, function, fragment, and other. The ``other'' category comprises words that are characteristic of conversational speech (vs. written text), including
words that function as backchannels (uh-huh, um-hum, huh, ...), filled pauses (um, uh), interjections (oh, ooh), and single word responses that can play the role of a backchannel (yeah, nope, huh, nah). 
Our hypothesis was that these categories would differ substantially in the tendency for transcribers to miss or hallucinate the words.

\begin{figure}[hptb]
\centering
\includegraphics[scale=0.4]{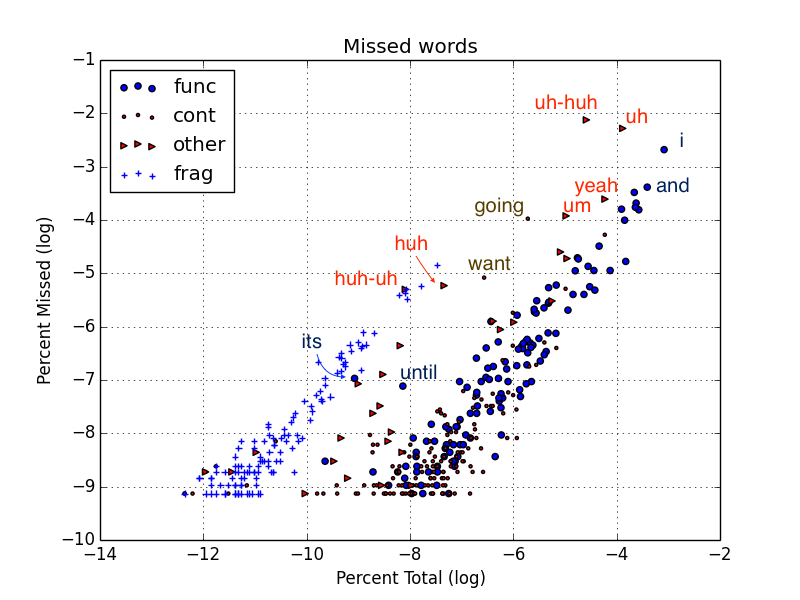}
\caption{Log relative frequency of missed instances of different words compared to their overall frequency in the corpus, distinguishing between function, content, fragment, and other.}
\label{fig:missed}
\end{figure}

\begin{figure}[hptb]
\centering
\includegraphics[scale=0.4]{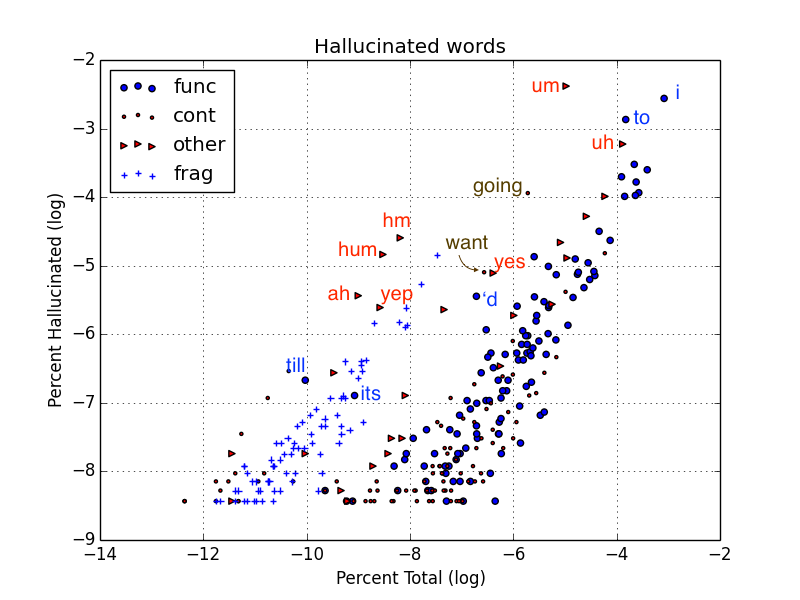}
\caption{Log relative frequency of hallucinated instances of different words compared to their overall frequency in the corpus, distinguishing between function, content, fragment, and other.}
\label{fig:hallucinated}
\end{figure}

Figures~\ref{fig:missed} and \ref{fig:hallucinated} show the log relative frequency of missed and hallucinated instances of each word, with different colors/symbols indicating the word category. 
For the most part, the relative frequency of a particular word being missed or hallucinated is proportional to the frequency of that word, but there is an offset that distinguishes different types. 
Word fragments are missed and hallucinated at a higher rate than other words of similar frequency. The high rate of fragment hallucinations is primarily due to transcription substitutions, e.g. `th-' vs.\ `thi-'. 
Function and content words follow a similar trend, with notable exceptions corresponding to transcription convention substitutions (`wanna' vs.\ `want to', `gonna' vs. `going to', `till' vs.\ `until', `its' vs.\ 'it's').


The other words as a group have atypically high frequencies, likely because many represent non-standard words.
In terms of transcription errors, the merger of substitutions into the missed/hallucinated categories obscures some differences. For example, `um' is infrequently missed and almost never hallucinated (unlike 'uh'), but often substituted.  
These differences are consistent with previous observations that the two filled pauses tend to be used differently by speakers.
It has also been observed that transcript errors on `um' and `uh' are substantially higher on average for those people who transcribed only a few conversations compared to those who had transcribed a large number \cite{le2017and}.
We hypothesize that the same will be true for disfluencies in speech that involve repetition or correction. In other words, 
listeners may not be consciously aware of many disfluencies even though they unconsciously use them in interpreting the intent of the interlocutor.  

\subsection{Disfluent Regions}

We hypothesized that transcribers would be more likely to both miss and mistranscribe words in the reparandum of a disfluency. In Table~\ref{tab:stats_mis_sub}, we provide the overall rate (relative frequency) of words associated with the different disfluency types $x$, together with the pointwise mutual information (PMI) of types and different error categories: PMI$(x,c) = \log [P(x|c)/P(x)]$. PMI greater than zero implies that the transcription error is more likely to occur in region $x$ than would be predicted by its overall relative frequency. Relative frequencies of words occurring in the reparandum of different disfluency types ($P_M$) are computed based on the MsState transcripts using the mapped disfluency labels. For most disfluency types, words that are completely missed (excluding substitutions) are even more likely to be in a disfluency reparandum.

\begin{table}
  \centering
  \begin{tabular}{|l||c|r|r|r|} \hline
   & & \multicolumn{3}{c|}{PMI$(x,y)$} \\ \cline{3-5}
    $x$ &  $P_M(x)$ & $(x,m*)$ & $(x,m)$ &  $(x,h)$ \\ \hline
    restart     & 0.003  & 0.49 & 0.64  & 0.60  \\ \hline
    repetition  & 0.032  & 0.61 & 0.41  & 0.22  \\ \hline
    repair      & 0.025  & 0.70 & 0.48  & 0.22  \\ \hline
    complex     & 0.003  & 0.61 & 0.40  & 0.54  \\ \hline
    fluent      & 0.879  & -0.19 & -0.08 & -0.01  \\ \hline
  \end{tabular}
  \vskip 0.1in
  \caption{Relative frequency of different disfluency types and the PMI associated with different reparandum word error categories ($m*$ = full miss, $m$ = miss, $h$ = hallucinate). 
  \label{tab:stats_mis_sub}}
\end{table}

The results show that transcription errors of all categories are more often associated with disfluent regions than with fluent speech, consistent with the overall rate of disfluencies being slightly higher in the MsState transcripts than in the Treebank transcripts (12.1\% vs. 10.8\%, respectively). 
As expected, the effect is stronger for missed words.
The high rate of errors in restarts may be indicative of these being less attended to by human listeners or a higher incidence of fragments here, but this is also the category where  inter-annotator agreement are least reliable.

While words seem to be more frequently misperceived in the reparandum of a disfluency, it may be that the disruption of disfluency is still perceived. For example, if the repetition {\it  I I I} is transcribed as {\it I I,} it is clear that the transcriber perceived a disfluency. From the analysis of interruption points in Sec.\ \ref{sec:quality}, we know that disfluencies are much more often missed than hallucinated. From analysis of the gold annotations, we find that both repetition and repair spans are missed at similar rates, but the hallucinations mostly involve repairs and restarts.


\section{Disfluency Detection Experiments}

The experiments in this section leverage a neural disfluency detection system described in \cite{zayats2018robust}. The model uses multiple levels to automatically find patterns in sentences. The first level  calculates similarities for each word in a sentence with words in the surrounding window, which we refer to as neighbor similarity. After calculating the single-token similarity weights, the second level use those weights as features to extract local patterns using a convolutional neural network followed by a max-pooling layer. We flatten the resulting outputs and concatenate with the word embeddings,
and input the resulting vector to a bidirectional LSTM-CRF.

We train two versions of the model (one using the original Switchboard transcripts and disfluency annotations, and the other using the corrected transcripts with silver mapped annotations) and assess performance on the corresponding two versions of the test set plus a subset that has been fully hand-corrected (gold). The results are presented in Table~\ref{tab:disfl_detect}. 

\begin{table}
    \centering
    \begin{tabular}{|cc|cc|}\hline
  &     & \multicolumn{2}{c|}{Training Data}\\
  Test set & Transcript &  Original & Corrected \\ \hline
   Full &    Original  & 87.80 & 87.23 \\
       & Silver & 88.67 & 86.96 \\ \hline
   Gold & Original & 88.69 & 88.54 \\
   Subset & Silver & 89.17 & 87.00 \\ 
         &  Gold   & 88.69 & 89.73 \\ \hline
    \end{tabular}
    \caption{Disfluency detection results training and testing on different versions of the annotations}
    \label{tab:disfl_detect}
\end{table}

The results on the original vs.\ silver annotations are not directly comparable because of the different definition of ground truth, but the comparison is useful for interpreting prior disfluency detection work that used the Treebank ground truth. The comparison of results on silver vs.\ gold transcripts provides an indication of the noise associated with the imperfect silver transcripts.
Lastly, we find that training on the original transcripts works well with both the original and silver test annotations, but training with the corrected silver transcripts gives the best result on the gold test set.




\section{Summary}

In summary, this study has shown that human transcribers tend to misperceive words proportionately to the frequency of those words, confirming the hypothesis that words which carry more information are less likely to be misperceived. Notable exceptions include words that are characteristic of spontaneous speech: filled pauses, interjections, and backchannels.
Disfluencies are also associated with higher rates of transcription errors. Listeners often miss repetitions, but they seem to perceive that there was a disfluency when there is a repair. 
To support this and future analyses, this work has provided a new version of Switchboard disfluency annotations. These annotations support more exploration of prosodic cues and disfluency detection \cite{zayats2019}.

This work was motivated in part by a prior study showing that transcription errors impact findings related to the usefulness of prosodic features in parsing \cite{Tran2018}, i.e., a significant fraction of the cases where prosody seems to hurt parsing are associated with transcription errors. 
The availability of the new disfluency annotations 
will make it possible to explore this question for disfluencies. 

For speech recognition applications, having high accuracy transcriptions does not seem to be critical. However, in spoken language processing and translation, disfluencies can impact performance. In addition, there are medical and educational applications where detected disfluencies may provide useful information about the speakers cognitive state. 



\section{Acknowledgements}
This work was funded in part by the US National Science Foundation, grant IIS-1617176.



\bibliographystyle{ieeetr}

\bibliography{refs,interspeech19,disfl}

\end{document}